\documentclass[runningheads]{llncs}
\usepackage[T1]{fontenc}
\usepackage{graphicx}
\usepackage{hyperref}
\usepackage{amsmath}
\usepackage{wrapfig}
\usepackage{orcidlink}
\begin{document}
%
%\title{Generating Robotics Arm Trajectories of Required Shape using Neural Networks} % Igor
\title{Generating and Customizing Robotic Arm Trajectories using Neural Networks} % Matilde
\titlerunning{Generating and Customizing Robotic Arm Trajectories}
\authorrunning{L\'u\v{c}ny A. et al.}
%\newcommand{\anonymize}[1]{\author{Anonymous submission}}
%\anonymize{
\author{Andrej L\'u\v{c}ny\inst{1}\orcidlink{0000-0001-6042-7434} \and
Matilde Antonj\inst{2,3}\orcidlink{0000-0003-2500-7754} \and \\
Carlo Mazzola\inst{3}\orcidlink{0000-0002-9282-9873} \and
Hana Horn\'a\v{c}kov\'a\inst{1}\orcidlink{0009-0006-3849-4159} \and
Igor Farka\v{s}\inst{1}\orcidlink{0000-0003-3503-2080}
%}
%
\authorrunning{A. L\'u\v{c}ny et al.}
\institute{Faculty of Mathematics, Physics and Informatics\\
Comenius University Bratislava, Slovakia\\
\email{\{lucny,hornackova52,farkas\}@fmph.uniba.sk}%\\
%\url{http://cogsci.fmph.uniba.sk/cnc/} 
\and
DIBRIS, University of Genoa, Genoa, Italy
\and
CONTACT, Italian Institute of Technology Genoa, Italy\\
\email{\{Carlo.Mazzola,Matilde.Antonj\}@iit.it}%\\
%\url{https://contact.iit.it/}
}
}
\maketitle              % typeset the header of the contribution
\begin{abstract}
We introduce a neural network approach for generating and customizing the trajectory of a robotic arm, that guarantees precision and repeatability. To highlight the potential of this novel method, we describe the design and implementation of the technique and show its application in an experimental setting of cognitive robotics. In this scenario, the NICO robot was characterized by the ability to point to specific points in space with precise linear movements, increasing the predictability of the robotic action during its interaction with humans. To achieve this goal, the neural network computes the forward kinematics of the robot arm. By integrating it with a generator of joint angles, another neural network was developed and trained on an artificial dataset created from suitable start and end poses of the robotic arm. Through the computation of angular velocities, the robot was characterized by its ability to perform the movement, and the quality of its action was evaluated in terms of shape and accuracy. Thanks to its broad applicability, our approach successfully generates precise trajectories that could be customized in their shape and adapted to different settings.
%We introduce an alternative approach to generating robotic arm trajectories constrained by a specific shape (inverse kinematics). For example, we need a robot pointing its finger to a given spot, approaching it, and finally touching it so we move its forefinger along a line. We design a neural network that performs forward kinematics of the robot arm. Then, we include it in a network, for which training on a suitably generated dataset provides poses of the robot (angles) at each trajectory instance. Finally, we calculate angular velocities, perform the movement, and evaluate its quality. 
%Using our approach, we have successfully generated trajectories for an experiment in cognitive robotics in which the investigation of the legibility of the robot movement solicited a particular and exact movement shape. Moreover, our method is general and can serve a variety of tasks with similar needs.
The code is released at 
%\url{https://github.com/andylucny/nico2}
\url{https://github.com/andylucny/nico2/tree/main/generate}.
% more specific

\end{abstract}
\section{Introduction}

In human--robot interaction (HRI) scenarios, the design of robotic motion must extend beyond efficiency in reaching a goal; it must also support human ability to interpret, predict, and feel safe around robotic actions. When humans and robots share the same physical space, the legibility of robot trajectories -- the ease with which a human observer can infer the robot intent \cite{ref_carlo1} -- becomes critical for effective, safe, and explainable collaboration. We can characterize legible movement by its distinctiveness, helping observers disambiguate between potential goals based on robot behavior \cite{ref_carlo2}.

To study human perception while interacting with robots, it is essential to design the robotic behavior to be repeatable and controllable \cite{ref_carlo3,ref_carlo4,ref_carlo5}. 
In this context, we address the problem of generating high-precise and controllable robotic arm trajectories that exhibit given shapes to serve as a foundation for studying human perception of robot legibility. Precision is crucial: subtle inaccuracies or inconsistencies in the generated trajectories could confound human interpretation, undermining the experimental study of legibility. Our work, therefore, aims to provide a method for creating repeatable, controlled motion patterns, enabling systematic investigation of how humans perceive and predict robot movements during embodied interaction. Theoretically, it should be possible to meet our needs with inverse kinematics (IK). However, we finally succeeded with a custom approach due to practical problems, including a lack of the robot model in a suitable format for the framework that matches our accommodated hardware, including the forefinger, combining movement with pointing. 

To this end, we propose a neural network-based approach for trajectory generation capable of producing smooth and accurate movements corresponding to predefined shapes. This method constitutes an important step toward developing repeatable and scalable tools to study motion legibility in HRI. Ultimately, it addresses the broader challenges of reducing robot unpredictability and improving interpretability and safety in shared environments \cite{ref_carlo6}.

In detail, implementing software for an experiment in cognitive robotics, we faced a problem in calculating the trajectories of a humanoid robot arm moving in line with a forefinger pointing in the direction of the movement to a specified point. This problem we can theoretically solve by any method of IK that we gradually apply for iterative calculation of the arm angles for each point (in 3D) of the line starting from the previous such point and for the orientation vector (roll, pitch, yaw) that is constant along the whole line and corresponds to its direction. 
%Though we had several ready-made solutions, for several reasons, mainly related to their reliability and accuracy, we finally had to develop a custom one. 
Since we aimed to simplify our job as much as possible, we employed a powerful library dedicated to training neural networks (Pytorch). 
%In this way, we have designed an alternative method that we consider interesting enough to be presented here.

Forward and inverse kinematics are fundamental concepts in robotics. Forward kinematics (FK) is the computation of the end-effector position and orientation based on known joint parameters, typically using Denavit-Hartenberg convention variant \cite{ref_dh1}. In contrast, IK addresses the problem of determining the joint parameters required to achieve the target position and orientation of the end effector \cite{ref_dh3}. This problem is often more complex due to the possibility of multiple solutions, redundancy, and the necessity of holding some given constraints, such as ranges of joint angles. 
In our case, we also need to follow a given trajectory shape. Analytical solutions exist for simple manipulators, but we must employ iterative methods for more complicated effectors like our robotic arm \cite{ref_dh4}. 

Therefore, we have designed the method described in detail below. It benefits from the power of today's tools for training neural networks, is easy to use, and is suitable for controlling the movement of robots along any chosen trajectory. It does not suffer from constraints like the FABRIK algorithm \cite{ref_fabrik,ref_fabrik_modified}, and unlike most traditional approaches, it generates a complete trajectory in one step.

Unlike recent approaches based on solving the IK problem using neural networks 
\cite{ref_iknn1,ref_iknn2,ref_iknn3}, we do not need to leverage large datasets of robot poses and corresponding joint configurations, which is a typical condition for the application of machine learning. In contrast, our method can provide such data. The feedforward neural networks, which are trained to predict joint configurations given a desired position of the end effector, can stem from them. Our method contributes to these approaches since deep learning models solving the inverse kinematics are often trained on synthetic data generated by robot simulation environments.
Our method is general and does not depend on a particular type of robot. On the other hand, all the concrete values we employ in our examples are relevant to our humanoid robot NICO \cite{ref_nico}, developed at the University of Hamburg and slightly improved for our purposes.

\section{Method}

First, we implement forward kinematics for our robot, which enables us to calculate position and orientation from angles and gradients of angles from the gradients of position and orientation. Then, we incorporate this module into a network, in which training provides us with IK constrained by ranges of angles. Finally, we run this network in parallel for all trajectory points and regularize the poses based on their training so that they not only solve IK for individual trajectory points but also correspond to fluent overall movement.

\subsection{Formal definition of the problem}

\newcommand{\vv}[1]{\boldsymbol#1}
Our method requires measuring the start and end poses of the robotic arm (given by joint angles), the shape of the trajectory, and the number of steps to control the motors. The start pose $\vv{p^{\rm s}}=(\theta^{\rm s}_1,\theta^{\rm s}_2,\dots,\theta^{\rm s}_m)$ and the end pose $\vv{p^{\rm e}}=(\theta^{\rm e}_1,\theta^{\rm e}_2,\dots,\theta^{\rm e}_m)$ are tuples of angles; in our particular case, we use $m=7$ degrees of freedom.
From these poses, we calculate the goal start point $\vv{P^{\rm s}} = (x_{\rm s}, y_{\rm s}, z_{\rm s})$ and the goal end point $\vv{P^{\rm e}} = (x_{\rm e}, y_{\rm e}, z_{\rm e})$ by FK described below. We translate the trajectory of the desired shape with these points and divide it into $n$ segments. This division must be accurate enough to assume constant angular velocities for individual motors within one segment. 
However, the duration of these segments must correspond to our ability to poll the motors via the bus they are connected to. We have to choose $n$ to balance the two requirements. 

The division provides us with $n+1$ goal points $\vv{P^{\rm g}_i}$, where $\vv{P^{\rm g}_0}=\vv{P^{\rm s}}$ and $\vv{P^{\rm g}_n}=\vv{P^{\rm e}}$. When the requested shape is a straight line, the goal points are
\begin{equation}
\vv{P^{\rm g}_i} = (x_{\rm s}+\frac{i}{n}(x_{\rm e}-x_{\rm s}), y_{\rm s}+\frac{i}{n}(y_{\rm e}-y_{\rm s}), z_{\rm s}+\frac{i}{n}(z_{\rm e}-z_{\rm s})) \text{~for~} i=0,1,\dots, n 
\end{equation}

However, when the shape is different, we can implement any strategy for their generation, e.g. splines.

In addition, we can derive $n$ goal vectors $\vv{v^{\rm g}_i}$ from the $n+1$ goal points as $\vv{v^{\rm g}_i} = \vv{P^{\rm g}_{i+1}} - \vv{P^{\rm g}_i}$ for $i=0,1,\dots,n-1$. All vectors are identical when the trajectory shape is a line; in this case, $\vv{v_i} = (\vv{P^{\rm e}}-\vv{P^{\rm s}})/n$.
The method provides us with $n+1$ poses of the robot arm (joint angles) $\{ (\theta^i_1,\theta^i_2,\dots \theta^i_m \mid i=0,1,\dots, n \}$. From them, we can easily calculate the corresponding angular velocities that move all the motors during the whole segment, taking $T/n$ ms, where $T$ is the selected duration of the whole movement in milliseconds.

\subsection{Neural network architecture for forward kinematics}

We build the FK module from four building blocks:
\[
\begin{alignedat}{2}
T(x,y,z)     &\;= 
\begin{bmatrix}
~1~~ & ~0~~ & ~0~~ & ~t_x \\
0 & 1 & 0 & t_y \\
0 & 0 & 1 & t_z \\
0 & 0 & 0 & 1
\end{bmatrix}
&\qquad
R_x(\theta) &\;= 
\begin{bmatrix}
1 & 0 & 0 & 0 \\
0 & \cos\theta & -\sin\theta & 0 \\
0 & \sin\theta & \cos\theta & 0 \\
0 & 0 & 0 & 1
\end{bmatrix} \\[1em]
R_y(\theta) &\;= 
\begin{bmatrix}
\cos\theta & 0 & -\sin\theta & 0 \\
0 & 1 & 0 & 0 \\
\sin\theta & 0 & \cos\theta & 0 \\
0 & 0 & 0 & 1
\end{bmatrix}
&\qquad
R_z(\theta) &\;= 
\begin{bmatrix}
\cos\theta & -\sin\theta & 0 & 0 \\
\sin\theta & \cos\theta & 0 & 0 \\
0 & 0 & 1 & 0 \\
0 & 0 & 0 & 1
\end{bmatrix}
\end{alignedat}
\]
where $T(x, y, z)$ is translation in 3D space,  $R_x(\theta), R_y(\theta), R_z(\theta)$ are rotations given by $\theta$ around $x,y,z$ axes, respectively.

\newcommand{\matmul}{\times} 
We can express FK as a product $M_0 \matmul M_1 \matmul \dots \matmul M_{k-1}$ of such matrices that correspond to our robot mechanics \cite{ref_dh2,ref_dk}. When we project the vector $(0,0,0,1)$ with these transforms, we gradually get points for individual joints in the form of $(x,y,z,1)$; the last of them being the point of the end effector:
\begin{equation}\label{eq2}
{(x, y, z, 1)}^\top = M_{k-1} \matmul \dots \matmul   M_1 \matmul M_0 \matmul {(0,0,0,1)}^\top
\end{equation}
Similarly, when using $3\times 3$ parts of the rotational matrices, we can project the vector $(0,0,1)$ to get joint orientations, including the roll, pitch, and yaw of the end effector.
\begin{equation}
{(\text{roll},\text{pitch},\text{yaw})}^\top = R_{l-1} \matmul \dots \matmul R_1 \matmul R_0 \matmul {(0,0,1)}^\top
\end{equation}
The forward kinematics of the right arm of NICO robot corresponds to
%\vspace{-5pt}
\begin{align}
\text{FK}&(\theta_1,\theta_2,\theta_3,\theta_4,\theta_5,\theta_6,\theta_7) = \label{eqdk1} \\ & \, T(0,5,19.5) \matmul R_z(90^\circ) \matmul R_z(\theta_1) \\ & \matmul
        T(0,-1.5,2.5) \matmul R_y(90^\circ)\matmul R_z(\theta_2) \\ & \matmul
        T(3,0,9.5) \matmul R_x(-90^\circ) \matmul R_z(-\theta_3) \\ & \matmul
        T(17.5,0,0) \matmul R_x(90^\circ) \matmul R_z(180^\circ) \matmul R_z(-\theta_4) \\ & \matmul
        T(10,0,0) \matmul R_y(90^\circ) \matmul R_z(-\theta_5/2) \\ &\matmul
        T(0,0,10) \matmul R_x(-90^\circ) \matmul R_z(-90^\circ) \matmul R_z(\theta_6/4.5+10) \\
       & \matmul T(0,-1,0) \matmul T(6,0,0) \matmul R_z(20+(\theta_7+180)/4.5) \\
       & \matmul T(6,0,0) \matmul R_y(90^\circ) \label{eqdk2}
\end{align}

The reference system has its origin $(0, 0, 0)$ at the base of the robot torso. The $x$-axis points behind the robot, the $y$-axis points laterally to its right, and the $z$-axis is parallel to gravity but pointing upward. Each line above moves the point to the next joint: the first line to \textit{shoulder\_z}, the second to \textit{shoulder\_y}, the third to \textit{arm\_x}, and the next ones to \textit{elbow\_y}, \textit{wrist\_z}, \textit{wrist\_x}, and the last one to \textit{indexfinger\_x}.
(Angles for the wrist and fingers are measured only in degree-like units, so we must recalculate them to degrees.)

Unlike the traditional approach, we never evaluate the product of all matrices 
$M_{k-1} \matmul \dots \matmul   M_1 \matmul M_0$ (see Eq. \ref{eq2}).
We always multiply the initial vector ${(0,0,0,1)}^\top$ by matrices one by one:
$(M_{k-1} \matmul \dots \matmul   (M_1 \matmul (M_0 \matmul {(0,0,0,1)}^\top)) \dots )$. 
The most important deviation is that the input to this module is not the initial vector but the joint angles $\theta_j$. We apply cosine and sine on them, update the matrices, and output the calculated position and orientation of the end effector. So, though the matrices linearly process vectors, this module is made to be a very nonlinear module of a neural network. Neither parameter of this module is trainable (although it could be within another project). However, we can run it not only in the forward mode that calculates point and orientation from angles but also in the backward mode that calculates angles' gradients upon the deviation of the point and orientation from their goal values. Also, it is essential to implement this module to receive not only a single input but also a batch, as is usual for neural networks.

\subsection{Neural network architecture for inverse kinematics}

With the module mentioned above, we can turn the IK problem (finding joint angles for a given 3D point and orientation) into training a neural network consisting of two parts: a generator of angles and the FK module. The generator provides proposals for angles, and the FK transforms them into the corresponding 3D point and orientation. Then, we aim to adjust the generator parameters to decrease the distance between the predicted values and the target.

Since we seek to train a neural network, we must formally define a dataset, while we know only one desired output (the given point and orientation). We can treat this requirement by assigning a formal input to the generator, always $1$. We process it by a linear layer of $m$ neurons, each having one weight and no bias, whose weights code the angles.

The primary design challenge for the generator is to operationalise the weights training while preserving the angle ranges. We need to ensure that the angles are within their range, can reach the extreme value, and break away from it as required during training. Clipping the angle values is not a solution due to zero gradients at the extremes. Therefore, first, we generate angles in the form of logits $z_j \in (-\infty, \infty)$, and then we apply a sigmoid and turn them to degrees, considering their range:
\begin{equation}
\theta_j = \theta_j^{\rm min} + \sigma(z_j) (\theta_j^{\rm max}-\theta_j^{\rm min}) 
\end{equation}

If we create a dataset with a single sample of input $1$ and output $x$, $y$, $z$, roll, pitch, and yaw, we could train the model until the weights in the linear layer correspond to the proper logits that provide the desired output. Then, we can use the above formula to get the corresponding values in degrees that form the required pose of the robotic arm.

\subsection{Neural network architecture for trajectory generation}

\begin{figure}
%\includegraphics[width=\textwidth]{schema.eps}
%\includesvg[width=0.2\textwidth]{schema.svg}
\includegraphics[width=\textwidth]{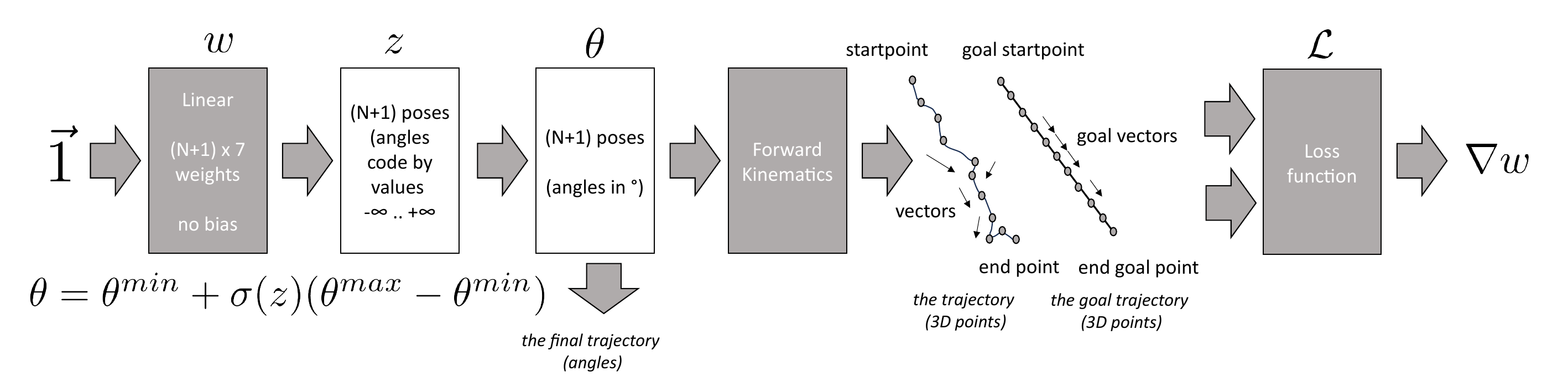} 
\caption{Schema of the neural network whose training leads to the generation of the robotic arm trajectory.} \label{fig:schema}
\end{figure}

If we use this kind of IK for all the goal points, the poses we get will implement a movement that is not fluent enough. We must solicit a global constraint to ensure the generated movement's fluency and make training of all poses in parallel. Nothing is easier within frameworks for training neural networks. We can express the fluency of the generated movement by minimizing angle differences through the generated poses. Implementing parallelization is also simple if the FK module already supports batches. We extend the number of inputs to the linear layer from one to $n+1$ and bypass the summation in its $neurons$, so its output has the shape $(n+1) \times m$. Then, we apply the same formula to keep angles in ranges to the tensor, which turns logits to angles in degrees $p_i$. In addition, we feed them as a batch into the FK module and get $(n+1)$ points $P_i$. Finally, we apply a suitable loss function compounded from individual IK problems for each point and the global constraint for the fluency of the trajectory (Figure~\ref{fig:schema}). 

Moreover, we prefer to match the start and end poses as precisely as possible. It is a good practice because these are the only two poses that we specified but also because the purpose of the trajectories can require their exact matching. 

As a result, our loss function will combine the following:
\newcommand{\norm}[1]{\lVert #1 \rVert}
\begin{itemize}

\item the mean distance between output points $\vv{P_i}$ and goal points $\vv{P^{\rm g}_i}$: \\
$\mathcal{L}_0 = \frac{1}{3n+3}\sum\limits_{i=0}^n \norm{\vv{P_i} - \vv{P^{\rm g}_i}}^2$

\item negative mean of cosines of angles between output orientation vectors $\vv{v_i}$ and goal vectors $\vv{v^{\rm g}_i}$ (in our particular case of the linear shape $\vv{v^{\rm g}_i} = (\vv{P^{\rm e}}-\vv{P^{\rm s}})/n$): %\\
$\mathcal{L}_1 = 1 - \frac{1}{n+1} \sum\limits_{i=0}^{n} {\frac{\vv{v_i}^\top \vv{v^{\rm g}_i}}{\norm{\vv{v_i}}\norm{\vv{v^{\rm g}_i}}}} $ 

\item distance between the start pose $\vv{p^{\rm s}}$ and its prediction $\vv{p_0}$: 
$\mathcal{L}_2 = \norm{\vv{p^{\rm s}} - \vv{p_0}}^2$

\item distance between the end pose $\vv{p^{\rm e}}$ and its prediction $\vv{p_n}$: 
$\mathcal{L}_3 = \norm{\vv{p^{\rm e}} - \vv{p_n}}^2 $

\item distance between the start point $\vv{P^{\rm s}}$ and its prediction $\vv{P_0}$: 
$\mathcal{L}_4 = \norm{\vv{P^{\rm s}} - \vv{P_0}}^2$

\item distance between the end point $\vv{P^{\rm e}}$ and its prediction $\vv{P_n}$: 
$\mathcal{L}_5 = \norm{\vv{P^{\rm e}} - \vv{P_n}}^2$

\item mean distance between the consecutive poses  
%\\ $\vv{p_i} = (\theta^i_1, \theta^i_2,\dots,\theta^i_m)$: 
%$\mathcal{L}_6 = \frac{1}{n m} \sum\limits_{i=0}^{n-1} \sum\limits_{j=1}%^{m} \norm{\theta^{i+1}_j -\theta^i_j}^2$
$\mathcal{L}_6 = \frac{1}{n m} \sum\limits_{i=0}^{n-1} \norm{\vv{p}_{i+1} - \vv{p}_i}^2$

\end{itemize}

The overall loss is a weighted sum of these seven losses
\begin{equation}
\mathcal{L} = c_0 \mathcal{L}_0 + c_1 \mathcal{L}_1 + c_2 \mathcal{L}_2 + c_3 \mathcal{L}_3 + c_4 \mathcal{L}_4 + c_5 \mathcal{L}_5 + c_6 \mathcal{L}_6
\end{equation}

Since we require very exact start and end poses, we put significant weights on the corresponding components of the loss function. Further, we equally weigh the influence of the position and orientation of the end effector. Finally, we assign a low but stable bias to the last component of the loss function that makes changes in angles as regular as possible. Based on experimentation, we have used weights: $c_0=1$, $c_1=50$,  $c_2=5$, $c_3=100$, $c_4=10$, $c_5=200$, $c_6=1$.

Generating the whole trajectory with one training process requires slightly modifying the dataset for the single IK problem. The dataset again contains a single sample. Its input is a vector of $n+1$ ones, and the desired output is the vector of the $n+1$ goal points and orientations. We use Adam optimizer with learning rate $0.1$. The initial setup of weights representing logits of angles follows analogical mixtures of the start and end poses as we mix the goal points from the start and endpoint. This training gradually achieves the required shape of the trajectory and improves its fluency. 
The training transforms an initial trajectory designed in the joint space of the robot to the trajectory of the required shape in the Cartesian space. Therefore, this method retrieves the 3D coordinates that the robot's end effector should assume to follow the required trajectory.
We can stop training when the angles do not change enough. We can also stop when the generated trajectory has a sufficient shape. The duration of training varies from seconds to minutes; it depends mainly on the end point and the availability of the sought trajectory. 

\section{Results}

%We have evaluated this method within the implementation of an experiment from the domain of cognitive robotics. This experiment investigates what a robot should do and doesn't have to do to make its movements human-readable. The robot moves its arm toward one of the target points on the touchscreen in various conditions, e.g., with or without a congruent gaze. Human participants must predict where the robot aims to touch while only partially performing its action. Therefore, we must generate a finite number of trajectories that end in the target points. The end pose of the trajectory must be exact since the robot touches a surface and must not move deeper. Also, the start pose must be precisely the same for all trajectories since we dislike giving the participant a cue on which trajectory will follow.
Our method of generating a trajectory for a robotic arm was evaluated to ensure the accuracy of the trajectories and to guarantee the required shape.
In the NICO robot we use its seven degrees of freedom: three in the shoulder, one in the elbow, two in the wrist, and one for the forefinger.
We divided the trajectory into $n = 50$ segments. As mentioned above, this number is a trade-off between considering the angular velocities constant within one segment (the larger $n$, the better) and polling motors (the lower $n$, the better) via the bus to which they are connected. In the case of our application, we connected the motors of the NICO robot via RS485, and it was not easy to balance the two requirements. We had to bypass the more sophisticated Pypot control interface and turn off command confirmation on the motors via the low-level Dynamixel SDK protocol. Only in this way did we achieve $n=50$, corresponding to 40 ms per segment, polling per 2-4 ms, considering the total requested duration of the movement $T=2s$. We described FK by the product of 24 translational or rotational matrices depending on the seven angle values (degrees of freedom) (Eq.~\ref{eqdk1}-\ref{eqdk2}). 

\begin{wrapfigure}{r}{7cm}
\centering
\vspace*{-5mm}
%\includegraphics[width=0.6\textwidth]{trajectories.eps}
%\includesvg[width=0.6\textwidth]{trajectories.svg}
\includegraphics[width=0.6\textwidth]{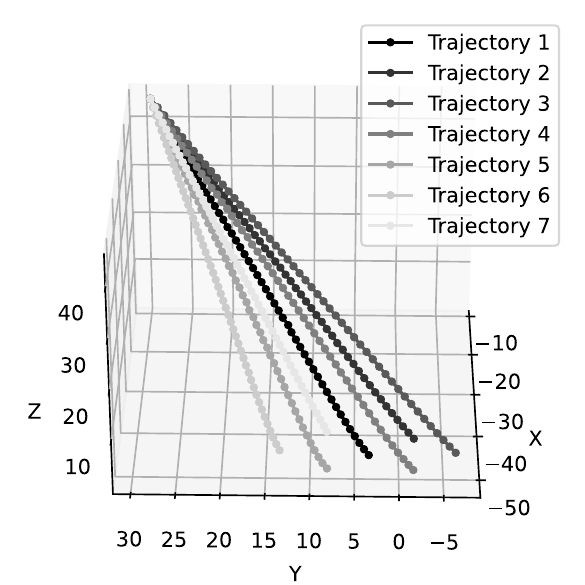}
\vspace*{-6mm}
\caption{The generated trajectories (front view). Each trajectory starts at the same start and ends at a different endpoint in the x-y plane. We can see that the trajectories have the required line shape.}
\vspace*{-4mm}
\label{fig:trajectories}
\end{wrapfigure}

Following the objectives of the experiment \cite{ref_kuz}, we generated seven trajectories to touch seven selected points on the touchscreen. They started from the same position with the raised arm and provided pointing by the forefinger to the point of touch throughout the movement, except in the starting phase. We recorded the start pose and seven touching poses. The NICO robot enables switching torque off and on and reading the current angles anytime, so managing this is quite comfortable. Training generated the trajectories successfully, as shown in Figure~\ref{fig:trajectories}. 
The number of iterations is provided in Table~\ref{tab:iters}. Since the NICO robot has only one degree of freedom at the elbow, it isn't easy to reach some spots with its arm, namely spots close to the robot. It may be the reason why closer points require a much higher number of iterations.

\begin{table}[ht]
\centering
\caption{Duration and accuracy of trajectories generation (our method)}
\begin{tabular}{|c|r|r|c|c|c|}
\hline
%trajectory id &  iterations & duration on CPU & loss & average distance from the line \\
\textbf{Trajectory} & \textbf{Iterations} & \textbf{Duration} & \textbf{Loss} & \textbf{Distance from} & \textbf{Pointing}\\ \textbf{ID} 
 & & \textbf{on CPU} & & \textbf{the line [mm]} & \textbf{deviation [$^\circ$]} \\
\hline
%CUDA is slower in our case (a small model and heavy communication between CPU and GPU
%1 & 621  & 13s & 8.7662 & & \\
%2 & 2429 & 58s & 9.8950 & & \\
%3 & 431  &  9s & 7.4980 & & \\
%4 & 1213 & 28s & 9.7828 & & \\
%5 & 2305 & 57s & 9.9352 & & \\
%6 & 388 & 9s & 8.5621 & & \\
%7 & 12037 & 326s & 9.9799 & & \\    
%
% mean distance has been fixed (mistake in the calculation)
% std added
% CPU
1 &     651~~ &   9 s~~~~ & ~~8.83~~ & $0.31 \pm 0.18$ & $12 \pm 11.8$\\
2 &   2374~~ & 29 s~~~~ & ~~9.82~~ & $0.26 \pm 0.13$ & $12 \pm 12.0$\\
3 &     374~~ &  5 s~~~~ & ~~7.51~~ & $0.28 \pm 0.13$ & $11 \pm 10.5$ \\
4 &   1153~~ & 15 s~~~~ & ~~9.98~~ & $0.38 \pm 0.22$ & $11 \pm 10.2$\\
5 &   2224~~ & 27 s~~~~ & ~~9.98~~ & $0.40 \pm 0.25$ & $12 \pm 11.3$\\
6 &     369~~ &   5 s~~~~ & ~~8.58~~ & $0.36 \pm 0.26$ & $13 \pm 13.1$\\
7 & 12394~~ & 151 s~~~~ & ~~9.96~~ & $0.23 \pm 0.23$ & $14 \pm 12.1$\\  
\hline
\end{tabular}
\label{tab:iters}
\end{table}

\begin{figure}[ht]
\centering
%\includegraphics[width=0.7\textwidth]{criterium2.eps}
%\includesvg[width=0.7\textwidth]{criterium2.svg}
\includegraphics[width=0.7\textwidth]{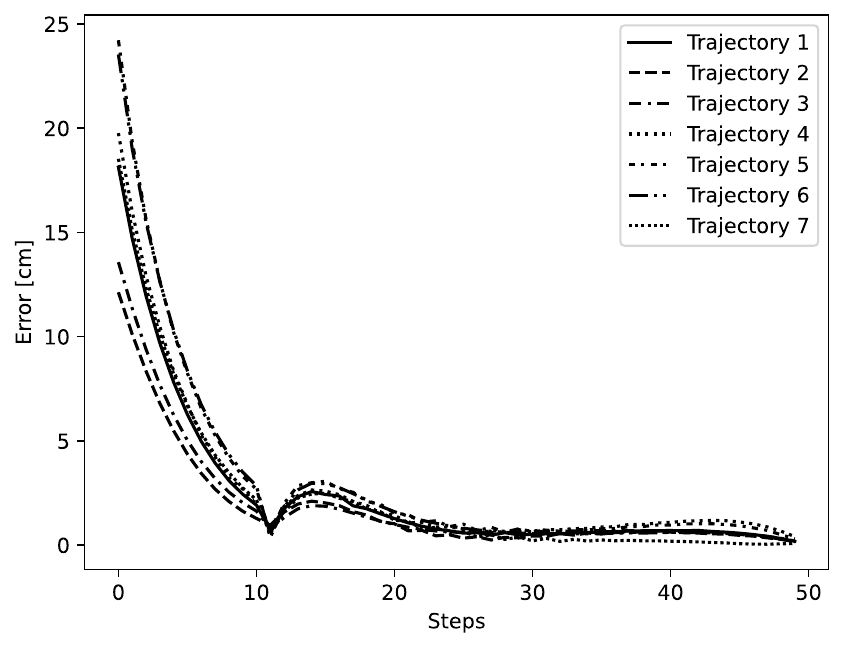}
\caption{The pointing error in cm for all trajectories (split into 50 steps). The closer the end effector is to the touching point, the more accurate the pointing is. From the beginning, it takes some time to point with the forefinger, and the fitting (from the last ten values) is less reliable for the first steps. The error has the character of damped oscillations, from which we see the absolute value.}
\label{fig:error}
\end{figure}

Evaluation of the quality of the generated trajectories employed the following criteria:
\begin{enumerate}
\item\label{criterium1}
The distance of the end effector point from the line and the pointing deviation (the last two columns in Table~\ref{tab:iters}).
\item\label{criterium2}
%The progress of the distance of the touching point from the intersection of the linear fit through points of moving end effector with the surface (where the robot points); 
The distance between the touching point on the surface and the intersection of the linear fit from the end effector to the surface. For each percentage of the trajectory, this distance was computed and evaluated 
(see Figure~\ref{fig:error}).
\end{enumerate}

While criterion~\ref{criterium1} focuses on the trajectory shape, criterion~\ref{criterium2} is more specific for our application related to the pointing action. The teaser video is available at \url{https://youtu.be/UIqqin3cJfs}.

Examining the impact of the individual parts of our loss function, we have performed the ablation study. Namely, avoiding losses $\mathcal{L}_0$ or $\mathcal{L}_6$ has a devastating effect on the whole movement, while the ablation of loss $\mathcal{L}_1$ moderately affects the pointing behavior. Ablation of $\mathcal{L}_0$ causes the generated trajectory to resemble the linear movement in the angle space; $\mathcal{L}_7$ causes the robot arm to shake.
The losses $\mathcal{L}_2$, $\mathcal{L}_3$, $\mathcal{L}_4$, and $\mathcal{L}_5$ are specific to our application, and we could avoid them for applications where exact start and end poses are not crucial. 

When we compare our method with traditional inverse kinematics, it is clear that we benefit from the ability to apply more local and global constraints to the trajectory. We have implemented a similar generation based on conventional inverse kinematics. We have selected the touching pose (more important than the start pose) as the initial pose, and we gradually called inverse kinematics for points on the line. We have specified not only the end-effector point, but also its orientation to be the same as the orientation of the line. In this case, trajectory generation is much faster; however, the accuracy of the pointing vector is much worse, and the start pose significantly varies (while it is constant by our method); see Table~\ref{tab:traditional}. The fluency of the movement is comparable. As a result, the traditionally generated trajectory is less legible. The teaser video is available at \url{https://youtu.be/PnguzMA5pDo}.

\begin{table}[ht]
\centering
\caption{Accuracy of trajectories generation (the traditional method)}
\begin{tabular}{|c|c|c|c|}
\hline
\textbf{Trajectory} & \textbf{Distance from} & \textbf{Pointing} & \textbf{Start point} \\ \textbf{ID} 
& \textbf{the line [mm]} & \textbf{deviation [$^\circ$]} & \textbf{variation [mm]} \\
\hline
1 & $2.84 \pm 1.73$ & $65 \pm 27.0$ & $5.6$ \\
2 & $3.38 \pm 2.09$ & $70 \pm 27.4$ & $5.9$ \\
3 & $1.39 \pm 0.54$ & $63 \pm 33.2$ & $6.9$ \\
4 & $3.02 \pm 1.87$ & $59 \pm 26.9$ & $5.6$ \\
5 & $2.91 \pm 1.66$ & $63 \pm 26.6$ & $5.4$\\
6 & $2.00 \pm 1.08$ & $63 \pm 30.2$ & $5.5$ \\
7 & $1.52 \pm 0.59$ & $70 \pm 33.6$ & $6.5$\\  
\hline
\end{tabular}
\label{tab:traditional}
\end{table}

To support the generality of our method, we have also applied it to another task requiring precise control of the robotic arm movement. We have implemented drawing letters in the air. We selected roughly a noncolliding position of the robot arm and projected the shape of a letter to the frontal plane passing through the end of the robot's forefinger. And train our neural network to get the trajectory. After 2000 iterations, we have come to a perfect solution. In this case, we can avoid losses for start and end positions since they are not crucial for this application, and all goal vectors will correspond to the normal vector of the drawing plane. The teaser video is available at \url{https://youtu.be/PsnmP7Kvx8g}.

\section{Conclusion}

This work introduced a novel, data-efficient method for generating robot arm trajectories constrained by desired start and end poses and a specified spatial path. Our approach is based on modern deep learning frameworks, leveraging a differentiable forward kinematics module integrated into a trainable neural network. This setup enables simultaneous inverse kinematics computation across the entire trajectory while maintaining motion fluency and respecting joint range constraints. Crucially, unlike typical learning-based approaches to IK, our method does not require pre-existing datasets. Instead, it could generate such data for a given set of trajectory shapes.

We demonstrated the utility of our method in a cognitive robotics experiment requiring human-interpretable robotic motion. Using the NICO humanoid robot, we successfully generated precise and smooth pointing trajectories, ensuring consistent start poses and exact endpoint accuracy, providing a new, efficient method of trajectory generation to customize and generate precise movements for the robot's arm.
%which were essential for the experimental design. 
The results confirmed both the practicality and generality of our approach, with trajectory generation converging within a reasonable time and the resulting motions fulfilling the task-specific requirements.

Our method presents an accessible and flexible alternative for trajectory generation in robotics. Its independence from large datasets, ability to enforce movement fluency, and ease of integration into standard neural network workflows make it a compelling tool for various applications, from human--robot interaction studies to general-purpose robotic control. Future work will include speeding up the method for real-time applications and integrating it with vision and reasoning models.

%\anonymize{
\begin{credits}
\subsubsection{\ackname} 
This work was supported by Horizon Europe project TERAIS, no. 101079338.
A.L. and I.F. were also supported by the Slovak Grant Agency for Science, project VEGA 1/0373/23.
\end{credits}
%}


\end{document}